\documentclass[runningheads]{llncs}
\usepackage{graphicx}
\usepackage{amssymb}
\usepackage{amsmath}
\usepackage{algorithm}
\usepackage{algpseudocode}

\usepackage[table,xcdraw]{xcolor}
\usepackage{multirow}
\usepackage{lipsum}

\begin{document}
\title{Barnes-Hut Approximation for Point Set Geodesic Shooting}
%
%
\author{Jiancong Wang$^*$  \inst{1} \and
Long Xie$^*$ \inst{2,3} \and
Paul Yushkevich\inst{3} \and
James Gee\inst{3}  }



%
%
\institute{Penn Image Computing and Science Laboratory, University of Pennsylvannia, PA 19104, USA \newline
\email{\{jiancong.wang,pauly2\}@pennmedicine.upenn.edu, long.xie@uphs.upenn.edu, gee@upenn.edu}}


%

\maketitle              
\begin{abstract}
Geodesic shooting has been successfully applied to diffeomorphic registration of point sets. Exact computation of the geodesic shooting between point sets, however, requires $O(N^2)$ calculations at each time step on the number of points in the point set. We propose an approximation approach based on the Barnes-Hut algorithm to speed up point set geodesic shooting. This approximation can reduce the algorithm complexity to $O(Nb+NlogN)$. The evaluation of the proposed method in both simulated images and the medial temporal lobe thickness analysis demonstrates a comparable accuracy to the exact point set geodesic shooting while offering up to 3-fold speed up. This improvement opens up a range of clinical research studies and practical problems to which the method can be effectively applied.
\end{abstract}

\section{Introduction}



In medical imaging, diffeomorphic registration is widely used to find a mapping from a biological structure to another since it ensures smoothness of the transformation and prevents folding of the manifold. In practice, the image-based computation of a diffeomorphic registration can be reduced to point matching of surfaces and curves \cite{joshi2000landmark} to minimize computational cost. Besides, the high-dimensional, dense time-varying velocity field in the registration can be represented as an initial momentum in a low-dimensional linear space, namely geodesic shooting \cite{allassonniere2005geodesic}. This representation of the deformation field allows a convenient characterization of subject variability by simple linear statistical methods, such as principal component analysis and principal geodesic analysis \cite{fletcher2003statistics}. 

Despite its favorable theoretical properties and relevance to statistical shape analysis, point set geodesic shooting has not been widely used in population studies or clinical applications because of the methods high computational complexity. Previous studies on surface matching under the diffeomorphic framework have produced valuable theoretical results and actual algorithms in applications. \cite{vaillant2005surface} builds a norm on the Hilbert space to represent surfaces as currents and derives a surface matching algorithm that preserves diffeomorphism. Extending this analysis under the Hilbert space, \cite{glaunes2008large} proposes a matching criterion to directly compare two curves with not-necessarily matched points. Previous works have attempted on minimizing computational cost. \cite{pai2016kernel} performs a Wendland kernel bundle stationary velocity field (wKB-SVF), and \cite{lombaert2013diffeomorphic} exploits an association graph to directly compare meshes.

In this work, we present an efficient algorithm on diffeomorphic point registration via geodesic shooting suitable for practical applications. Inspired by \cite{golyanik2016gravitational} and \cite{camassa2017geodesic}, our work reduces an $O(N^2)$ pairwise velocity computation step in geodesic shooting to two computationally light steps. The first step is an oct-tree construction step of all points by Barnes-Hut algorithm \cite{golyanik2016gravitational}, with time complexity $O(NlogN)$. The second step is a discontinuous conical Gaussian kernel approximation step, which requires a $O(Nb)$ time complexity, where $b$ is the average number of points within a 3$\sigma$ neighborhood. The proposed method is evaluated with both synthetic images and T1-MRI scans in an Alzheimer's disease(AD) medial temporal lobe (MTL) thickness study.

\section{Methods}

\subsection{Hamiltonian Formulation of Point Set Geodesic Shooting}

Let $X_p$ be the set of template points, $X_m$ the corresponding set of target points, and both sets have $N$ points. The algorithm determines a spatial transformation $\phi$ that optimally matches the points $X_p$ to $X_m$ under the diffeomorphic constraint. The objective function is a minimization of a deformation constraint term regularizing the transformation $||v||$ from time 0 to time 1 and a data attachment term minimizing the distance of the transformed points and the target. As proposed in \cite{allassonniere2005geodesic}, this function can be formulated as a minimization of a sum of kinetic energy and potential energy, and be solved under the Hamiltonian framework. Our notation is consistent with \cite{allassonniere2005geodesic} for clarity.

Let the positions $q$ of the points be a function of time $t \in [0, 1], q(t) = {q_1(t), . . . , q_N(t)}$ and the momentum of the points be $p(t) = {p_1(t), . . . , p_N(t)}$, $(p_i(t) \in \mathbb{R}^3)$. The initial positions of the points are given by $q(0) = X_j$, and the initial momenta of the points $p(0)$ are the unknowns that the geodesic shooting algorithm will determine. The evolution of the system is formulated in terms of the Hamiltonian $H(p, q) = < p, K(q)p >$, where $H(p, q)$ is the kinetic energy of the system and is constant over time and $K(q)$ is a $3N \times 3N$ with $N \times N$ diagonal blocks, with the $(i, n)$-th block equal to $G_{\sigma_{gs}} (|| q_i - q_m ||)·I_3$, $G_{\sigma_{gs}}$ being a Gaussian kernel and $I_3$ being an identity matrix. the evolution of the system is formulated in terms of the Hamiltonian:

\begin{equation}\label{A_Label}
\left \{
  \begin{split}
    \frac{dq}{dt} = \frac{\partial H}{ \partial p} (q, p), \\
    \frac{dp}{dt} = -\frac{\partial H}{ \partial q} (q, p)
  \end{split}
\right.
\end{equation}
The point matching problem is formulated as an optimization of
\begin{equation}
p_0^* = p_0 \in \mathbb{R}^{3L} H( q_0 , p_0 ) + \lambda || q(1) - X_m ||_2^2
\end{equation}

This minimization problem is discretized in time and solved using gradient-based optimization. In our implementation, we use L-BGFS \cite{liu1989limited} from VNL numeric library \cite{johnson2015itk} to perform quasi-second order update.

Given an initial momenta $p^*_{0}$ and the corresponding point trajectories $q^∗(t)$ , the point transformation can be interpolated over the entire spatial domain $x \in \Omega$ to yield a smooth velocity field

\begin{equation} \label{eq:3}
v(x, t) = \sum_{i=1}^{N} G_{\sigma_{gs}} (||x - q_i(t)||_2)p_i^*(t), x \in \mathbb{R}^{3}
\end{equation}

The velocity of each point is dependent on all other points with a Gaussian decay by equation \ref{eq:3}: $ v_j = \sum_{i=1, i \neq j}^{N} G_{\sigma_{gs}} (||q_j^* - q_i||_2)p_i, \forall j = 1...N$. This equation is used to update each point $q_j$, and to calculate its exact velocity. The total computation complexity of this procedure is $O(N^2)$ since it involves calculating the momentum of all other points weighted by the Gaussian distance kernel. 

\subsection{Barnes-Hut Approximation of Gaussian Kernel}

Based on the observation that the Gaussian of two distant points is close to 0, a grouping algorithm is proposed to assemble a cloud of points far away from a target point as a unit and use the average information of the group to avoid redundant calculation. This approximation algorithm is composed of two steps: 1) 3D spaces are recursively sub-divided into oct-trees and 2) each point is traversed down the trees and interaction with other tree nodes is calculated. Since the same procedure applies to each time step, the time variable $t$ is dropped in the following equations for simplicity.

\subsubsection{Oct-tree Representation of Space} 
An oct-tree of all points is constructed so that a node in a tree consisting of multiple points can be viewed as a whole to reduce calculation of all points. The oct-tree is constructed by subdividing the space recursively until each leaf node only contain a single point. First the global 3D bounding box of all points is found by $q_{\min/\max} = {\min/\max}{q \in q_i, i=1...N}q_i$. The root node is initialized to be $q_{min}$, $q_{max}$. Then points are added recursively to the tree and the tree structure gets updated. Suppose the $k-$th tree node $\mathcal{N}_k$ representing a cubic space is delimited by $q_{kmin}$, $q_{kmax}$, then each non-leave node contains eight child nodes, corresponding to the eight quadrant defined by bi-partition of the space between $q_{kmin}$ and $q_{kmax}$. 

Given a tree node $\mathcal{N}_k$ and a point $q_j$, the recursion policy is devided into three scenarios: (1) $\mathcal{N}_k$ is a non-leaf node, then $q_j$ is passed to the corresponding child node and recursion preceeds (2) $\mathcal{N}_k$ is an empty leaf node, then $q_j$ is added to $\mathcal{N}_k$ and the recursion stops, and (3) $\mathcal{N}_k$ is an occupied leaf node, then $\mathcal{N}_k$ is subdivided to eight quadrants and $q_j$, $q_i$ are passed to the corresponding child nodes. When $q_j$ is passed to node $\mathcal{N}_k$, some node statistics are updated accordingly: (1) the node's total momentum $p_{kc} \longleftarrow p_{kc} + p_j$, (2) number of points $n_k \longleftarrow n_{k} + 1$, (3) center of point position $q_{kc} \longleftarrow (q_{kc} \cdot n_k + q_j) / (q_j + 1)$, (4) actual range of the node $q_{kmax} \longleftarrow \max ( q_{kmax}, q_j  )$ and $q_{kmin} \longleftarrow \min ( q_{kmin}, q_j  )$. This information is a compact representation of the point subset in the node. 

\begin{figure}
\begin{center}
\includegraphics[width=10cm]{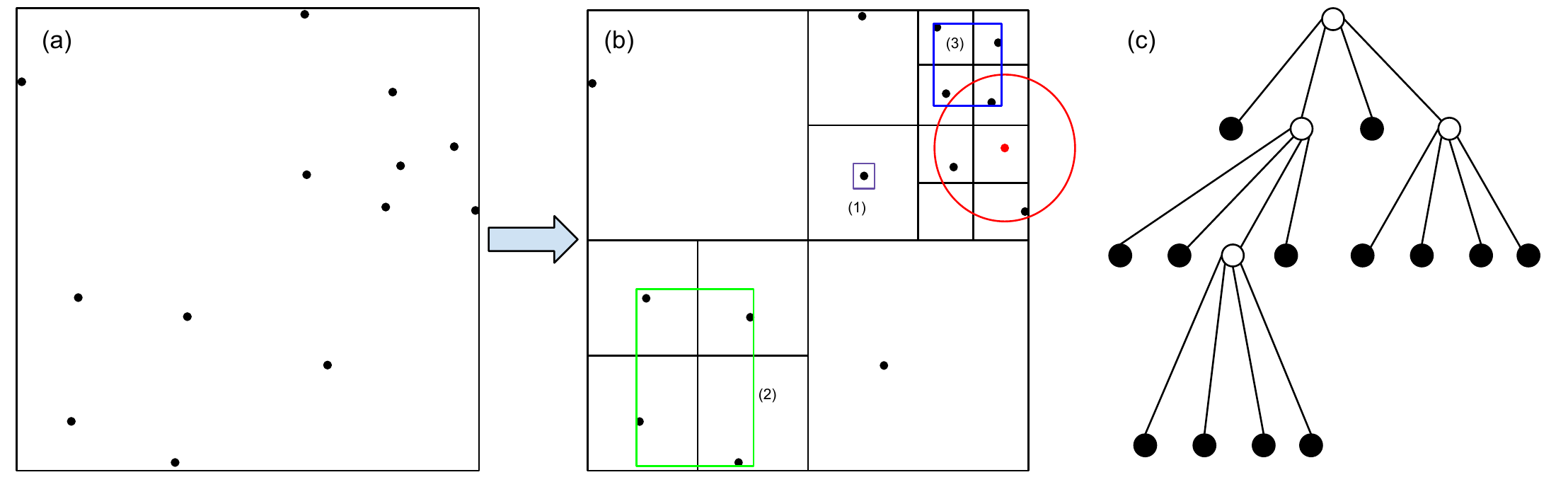}
\end{center}
\caption{2D graphic illustration of the recursive partition and Gaussian calculation. (a) Sample points. (b) Quad-tree subdivision. For a point in action (red), the Gaussian interaction (1) of a single-point node is directly calculated, (2) of a node outside the distance threshold (red circle) is approximated, (3) of a node with multiple points and at least one within the distance threshold is recursively calculated. (c) Quad-tree representation. The black and white nodes are respectively actual points and nodes.}
\label{fig1}
\end{figure}

In each step of the backward stage (gradient descent stage), the update of the point position derivative is inserted into the tree node $\mathcal{N}_k$ by $\alpha_{kc} \longleftarrow \alpha_{kc} + \alpha_i$ and momentum derivative $\beta_{kc} \longleftarrow \beta_{kc} + \beta_i$. The node statistics is modified in a similar way, including total point position derivative, total momentum derivative, number of points, center of point position, and range of node. The complexity of the tree building is of $O(N logN)$ in the best balanced oct-tree case.

\subsubsection{Tree-based Forward/Backward Gaussian Kernel Calculation}
The forward/backward pairwise Gaussian interaction is calculated based on the spatial oct-tree. For a point $q_j$ and a tree node $\mathcal{N}_k$, the recursion policy is again split into three scenarios in both forward and backward stages. (1) If $\mathcal{N}_k$ is a leaf node with only one point $q_i$, then the exact Gaussian kernel between $q_j$ and $q_i$ is calculated. (2) If $\mathcal{N}_k$ contains multiple points that are all farther than a distance threshold (3$\sigma$), then a single Gaussian interaction is calculated using the summarized total momentum $p_{kc}$ and center position $q_{kc}$. (3) If $\mathcal{N}_k$ contains multiple points but within the distance threshold, then the node gets subdivided. The tree traversal ideally has a complexity $ O(Nb + NlogN)$, where $b$ is the number of neighboring points within the distance threshold. In the experiment, we have $\sigma=2$, $k \approx 200$, $N \approx 2000$ and $b \approx 200$. Therefore, our scheme provides great speed up in theory. 

All accurate calculations can now be replaced with tree node approximations. In the forward stage, the actual Gaussian kernel weighted momentum is replaced with the approximation:
\begin{equation}
v_{jc} = \sum_{k=1}^{K_j} G_{\sigma_{gs}} (||q_j^* - q_{kc}||_2)p_{kc},
\end{equation}

where $K_j$ is the number of tree nodes traversed from $q_j$. The dot product between the exact momentum is also replaced with the average momentum, i.e., $p_{ji} = p_j \cdot p_i \longrightarrow p_{jk} = p_j \cdot p_{kc} / n_k$.

In the backward stage, besides the Gaussian kernel approximation and the momentum approximation which are the same as in the forward stage, the point position derivative and momentum derivative is replaced with their tree averaged counterparts, i.e., $\alpha_i \longrightarrow \alpha_{kc} / n_{k}$ and $\beta_i \longrightarrow \beta_{kc} / n_{k}$. 

\subsubsection{Implementation details} 
The algorithm is implemented in C++ with ITK 4.11 /VTK 7.1 libraries. Our code is compiled with gcc 4.4.7 and deployed on a cluster setting with E5-2643 v3 CPU, 16GB RAM, Centos 6 without multi-threading.

\section{Evaluation}
\subsection{Evaluation on Synthetic Data}
\begin{figure}

\begin{center}
\includegraphics[width = 12cm]{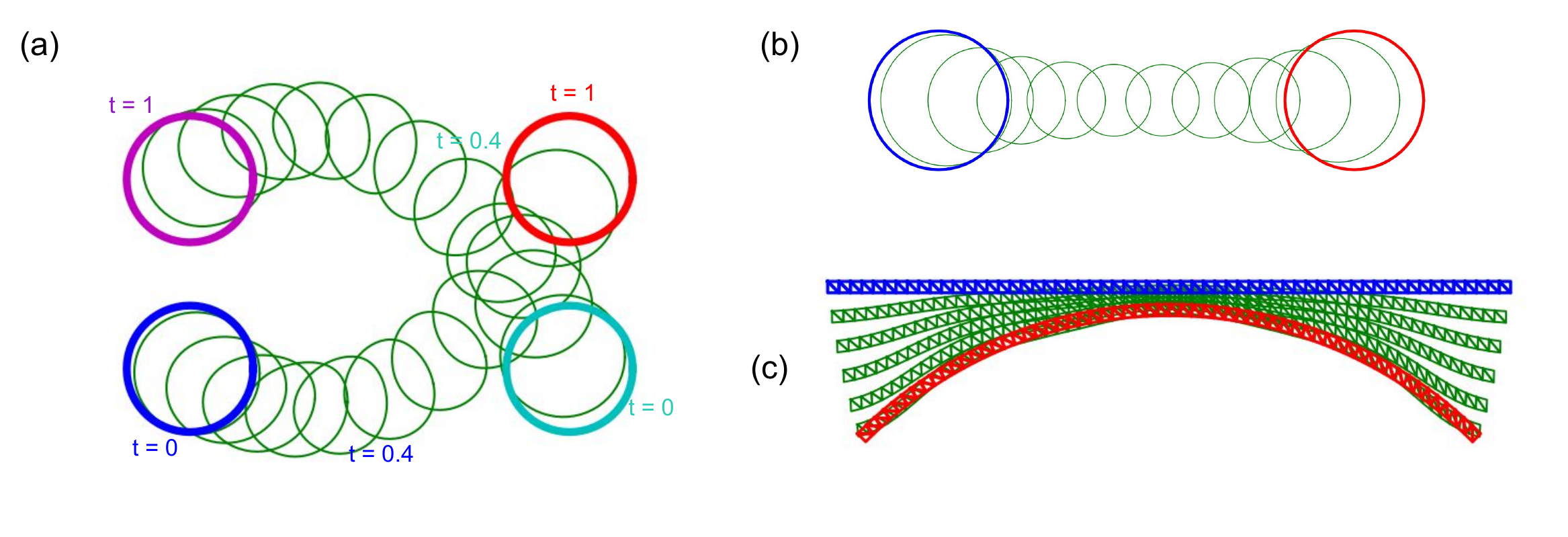}
\end{center}
\caption{Test of proposed algorithm on synthetic shapes. In (b) and (c), blue, red, and green colors are the moving shape, the fixed shape and the deformation respectively. In (a), the blue and green circles are registered to the red and purple circles simultaneously.} 
\label{fig_syntheticCases}
\end{figure}

The effectiveness of the proposed approximation is tested on (1) preservation of the diffeomorphic property (Fig. \ref{fig_syntheticCases}a, b) and (2) speed gain (Fig. \ref{fig_syntheticCases}c). In Fig. \ref{fig_syntheticCases}a, two circles (blue and green) are simultaneously registered to their counterparts (red and purple), while the two deformation fields do not collide or collapse. Synthetic case (b) registers the blue circle to the red. The shrinking-expanding artifact in the trace (green) is caused by the diffeomorphic constraint, the same as in \cite{arguillere2015shape}. Fig. \ref{fig_syntheticCases}c shows a registration of a long rectangular mesh (blue) to its bent counterpart (red), demonstrating a hypothetical use case where the Barnes-Hut approximation will speed up the registration significantly. Most points in this shape are far away from each other by 3$\sigma$, where Barnes-Hut approximation is useful. With a similar matching accuracy (evaluated as the sum of the square distance), our approximation achieves 2.15 folds speedup. (Table \ref{table_syntheticdata}). 
On the other hand, when points are relatively concentrated, such as in case (b), the Barnes-Hut approximation is much slower than the accurate version.

\begin{table}[]
\caption{Process time (min) and residual error (mm) of cases (b) and (c)}
\label{table_syntheticdata}
\centering
\begin{tabular}{|c|c|c|c|c|c|}
\hline
\multirow{2}{*}{\textbf{}} & \multirow{2}{*}{\textbf{\begin{tabular}[c]{@{}c@{}}Number of \\ points\end{tabular}}} & \multicolumn{2}{c|}{\textbf{BH-approx}} & \multicolumn{2}{c|}{\textbf{Non-approx}} \\ \cline{3-6} 
 &  & \textbf{Time} & \textbf{Residual error} & \textbf{Time} & \textbf{Residual error} \\ \hline
Circles (Fig. 2b) & 1200 & 124.6 & 3.4002e-5  & 40.4 & 1.5964e-5 \\ \hline
Flat shape (Fig. 2c) & 1202 & 23.0 & 0.1558 & 49.5 & 0.1507 \\ \hline
\end{tabular}
\end{table}

\subsection{Evaluation on Cortical Thickness Analysis of the Medial Temporal Lobe in Alzheimer's Disease}

To demonstrate the utility of the proposed technique in the real application, we apply the pipeline to perform thickness analysis for the medial temporal lobe (MTL) cortex in the context of early detection of AD. Since MTL is one of the earliest regions affected by neurofibilary tangle pathology (NFT), a biomarker directly linked to neuronal damage, the thickness measurement of the MTL subregions is a promising biomarker of AD.

\begin{figure}
\begin{center}
\includegraphics[width = 12cm]{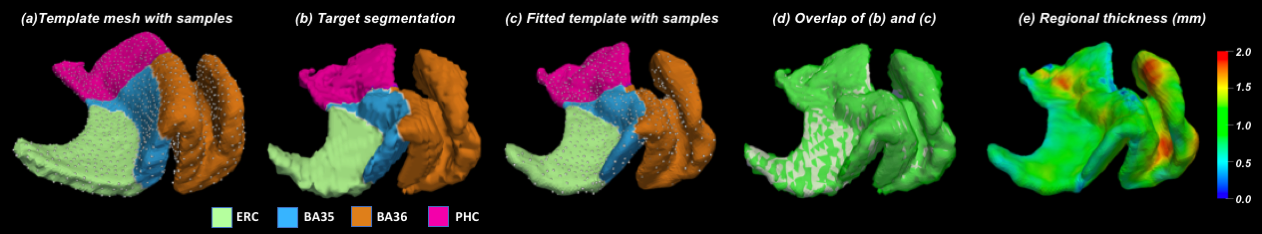}
\end{center}
\caption{An illustration of MTL thickness measurement pipeline} 
\label{MTLCortexFig}
\end{figure}

\subsubsection{Template Matching} 
Baseline T1-weighted MRI scans of 665 subjects from Alzheimer's disease neuroimaging initiative (ADNI) were chosen in this study. An open-source multi-atlas segmentation pipeline proposed by Xie et al. \cite{xie2016accounting} was applied to segment the MTL cortex subregions including the entorhinal cortex (ERC), Brodmann areas 35 and 36 (BA35/36) and parahippocampal cortex (PHC). The greedy registration package \footnote{sites.google.com/view/greedyreg} was applied to register the automatic segmentation to the MTL cortex template \footnote{Publicly available at github.com/LongXie/Multitemplate-for-MTLCortex-T1} as described in \cite{Xie2018}. The template space consists of the MTL cortex labels and a dense surface mesh of the union of ERC, BA35, BA36, and PHC labels (Fig. \ref{MTLCortexFig}a), from which a set of $N$ points were uniformly sampled using Poisson Disk Sampling \cite{Corsini2012}. The points were then warped to the space of each subject. The alignment of the points in the subject space to the template space consists of an initial rigid registration step using the Procrustes algorithm \cite{Drydmard16}, and a geodesic shooting step. The dense template mesh and the template labels are then warped to the space of the subject segmentation (Fig. \ref{MTLCortexFig}c).  

\subsubsection{Cortical Thickness Measures and Statistical Analysis}
To extract regional thickness, the pruned Voronoi skeleton \cite{Ogniewicz1995} is first extracted from the target mesh and the distance between each vertex and the closest point on the skeleton is computed (Fig. \ref{MTLCortexFig}e). The median of the thickness measures of each label is then extracted as the summary thickness measure of each label. Bilateral thickness measure of the same label is averaged. The Dice similarity coefficient (DSC) of each MTL cortex label between the warped template and the automatic segmentation is computed, as a measure of the quality of fit. 



\begin{table}[]
\newcolumntype{L}[1]{>{\raggedright\arraybackslash}p{#1}}
\newcolumntype{C}[1]{>{\centering\arraybackslash}p{#1}}
\newcolumntype{R}[1]{>{\raggedleft\arraybackslash}p{#1}}
\caption{Matching accuracy and processing time.}
\label{tab_time_DSC}
\centering
\fontsize{8.5}{8.2}
\selectfont
\begin{tabular}{|C{3.45cm}|C{1.73cm}|C{1.71cm}|C{1.71cm}|C{1.71cm}|C{1.71cm}|}
\hline
\rowcolor[HTML]{000000} 
{\color[HTML]{FFFFFF} } & {\color[HTML]{FFFFFF} \textbf{Time(min)}} & {\color[HTML]{FFFFFF} \textbf{ERC DSC}} & {\color[HTML]{FFFFFF} \textbf{BA35 DSC}} & {\color[HTML]{FFFFFF} \textbf{BA36 DSC}} & {\color[HTML]{FFFFFF} \textbf{PHC DSC}} \\ \hline
\textbf{Non-approx (1275 pts)} & $23.2 \pm 4.7$ & $0.92 \pm 0.02$ & $0.92 \pm 0.02$ & $0.92 \pm 0.02$ & $0.93 \pm 0.01$ \\ \hline
\textbf{BH-approx (1275 pts)} & $7.7 \pm 0.7$ & $0.92 \pm 0.02$ & $0.92 \pm 0.02$ & $0.92 \pm 0.02$ & $0.93 \pm 0.01$ \\ \hline
\textbf{BH-approx (2540 pts)} & $24.5 \pm 2.2$ & $0.93 \pm 0.02$ & $0.93 \pm 0.02$ & $0.93 \pm 0.02$ & $0.94 \pm 0.01$ \\ \hline
\end{tabular}
\end{table}

\subsubsection{Discussion}
The importance of faster implementation of point set geodesic shooting was motivated by the need to register data sets that are either very densely sampled or large in scale -- or in many circumstances both, for example, the ADNI study of MTL. The performance of the Barnes-Hut approximation and the original geodesic point shooting are compared on the same point data set to evaluate the speed gain and point matching accuracy. The number of points is set to $N = 1275$ so that the procession time of the entire dataset ($665$ subjects $\times 2$ hemispheres) is reasonable. As can be seen in Table~\ref{tab_time_DSC}, the Barnes-Hut approximation achieves approximately three-fold speedup compared to the original implementation with almost the same DSC for all the labels. 

We hypothesize that both the matching accuracy and discriminative power can be improved when processing with points sampled at a higher resolution. The repeated experiment with twice the sample density, i.e., 2540 points, shows a similar processing time (24.5 min) to the 1275 points experiment (23.2 min) but a higher DSC (Table~\ref{tab_time_DSC}). This finding can potentially be attributed to a more accurate thickness measurement with a higher resolution point set sampling of the imaged anatomy.


\section{Conclusion}
In this work, we present an approximation scheme based on the Barnes-Hut algorithm to perform geodesic point shooting, which can have a three-fold speed up without compromising matching accuracy. The proposed method allows calculation based on more densely sampled data sets, which in turn can translate to more accurate registrations and in turn more sensitive imaging-based markers.


%
%
%
%

\bibliographystyle{splncs04}
\bibliography{mybib.bib}

\end{document}


\begin{table}[]
\newcolumntype{L}[1]{>{\raggedright\arraybackslash}p{#1}}
\newcolumntype{C}[1]{>{\centering\arraybackslash}p{#1}}
\newcolumntype{R}[1]{>{\raggedleft\arraybackslash}p{#1}}
\centering
\caption{Characteristics of the ADNI dataset. All statistics are in comparison to A$\beta$- CN. Independent two-sample t-tests (age, education, MMSE) and contingency χ2 test (gender) were used. Standard deviation in parentheses. MMSE = mini-mental state examination. $^{*}: p < 0.05; ^{**}: p < 0.01; ^{***}: p < 0.001$.}
\label{tabADNI}
\fontsize{8}{8.2}
\selectfont
\begin{tabular}{|C{1.9cm}|C{1.9cm}|C{2cm}|C{2cm}|C{2cm}|C{1.9cm}|}
\hline
\rowcolor[HTML]{000000} 
{\color[HTML]{FFFFFF} } & {\color[HTML]{FFFFFF} \textbf{A$\beta$- CN}} & {\color[HTML]{FFFFFF} \textbf{A$\beta$+ CN}} & {\color[HTML]{FFFFFF} \textbf{A$\beta$+ EMCI}} & {\color[HTML]{FFFFFF} \textbf{A$\beta$+ LMCI}} & {\color[HTML]{FFFFFF} \textbf{A$\beta$+ AD}} \\ \hline
\textbf{N} & 190 & 95 & 142 & 110 & 128 \\ \hline
\textbf{Age (yrs.)} & 72.3 (6.0) & 74.8 (5.9)$^{***}$ & 73.6 (6.9) & 72.4 (6.8) & 74.3 (8.2)$^{*}$ \\ \hline
\textbf{Sex (M/F)} & 100/90 & 31/64$^{**}$ & 81/61 & 58/52 & 69/59 \\ \hline
\textbf{Edu (yrs.)} & 16.9 (2.4) & 16.1 (2.7)$^{*}$ & 15.6 (2.8)$^{***}$ & 16.6 (2.6) & 15.6 (2.7)$^{***}$ \\ \hline
\textbf{MMSE} & 29.0 (1.3) & 29.0 (1.1) & 28.0 (1.7)$^{***}$ & 27.2 (1.9)$^{***}$ & 23.0 (2.1)$^{***}$ \\ \hline
\end{tabular}
\end{table}

\begin{table}[]
\newcolumntype{L}[1]{>{\raggedright\arraybackslash}p{#1}}
\newcolumntype{C}[1]{>{\centering\arraybackslash}p{#1}}
\newcolumntype{R}[1]{>{\raggedleft\arraybackslash}p{#1}}
\caption{Statistical analysis results using thickness measurements extracted by the none-approximation (Non-approx), Barnes-Hut approximation (BH-approx) implementations, adjusted for age, in discriminating patient groups from normal controls in ADNI. Each patient group is compared to A$\beta$- CN separately. Measurements that survived Bonferroni correction ($p<0.05/12$) are highlighted in bold.}
\label{table3}
\centering
\fontsize{8}{8.2}
\selectfont
\begin{tabular}{L{1.75cm}|R{1.7cm}|R{1.55cm}|R{1.6cm}|R{1.71cm}|R{1.7cm}|R{1.6cm}}
\hline
\rowcolor[HTML]{000000} 
{\color[HTML]{FFFFFF} \textbf{}} & {\color[HTML]{FFFFFF} \textbf{}} & {\color[HTML]{FFFFFF} \textbf{A$\beta-$ CN}} & {\color[HTML]{FFFFFF} \textbf{A$\beta+$ CN}} & {\color[HTML]{FFFFFF} \textbf{A$\beta+$ EMCI}} & {\color[HTML]{FFFFFF} \textbf{A$\beta+$ LMCI}} & {\color[HTML]{FFFFFF} \textbf{A$\beta+$ AD}} \\ \hline
\textbf{N} &  & 190 & 95 & 142 & 110 & 128 \\ \hline
\rowcolor[HTML]{C0C0C0} 
\multicolumn{7}{l}{\cellcolor[HTML]{C0C0C0}\textbf{ERC thickness (mm), age as covariate}} \\ \hline
 & Median (SD) & 2.00 (0.16) & 2.00 (0.16) & 1.98 (0.17) & \textbf{1.92 (0.18)} & \textbf{1.78 (0.22)} \\ 
 & \textit{F} stats &  & \textless{}2.5 & \textless{}2.5 & \textbf{15.8} & \textbf{116.1} \\ 
\multirow{-3}{*}{\textbf{\begin{tabular}[c]{@{}l@{}}Non-approx \\ (1275 pts)\end{tabular}}} & \textit{p} value &  & \textgreater{}0.1 & \textgreater{}0.1 & \textbf{8.90e-05} & \textbf{3.00e-23} \\ \hline
\rowcolor[HTML]{EFEFEF} 
\cellcolor[HTML]{EFEFEF} & Median (SD) & 2.00 (0.15) & 2.00 (0.16) & 1.98 (0.17) & \textbf{1.92 (0.18)} & \textbf{1.78 (0.21)} \\ 
\rowcolor[HTML]{EFEFEF} 
\cellcolor[HTML]{EFEFEF} & \textit{F} stats &  & \textless{}2.5 & \textless{}2.5 & \textbf{16.3} & \textbf{116.5} \\ 
\rowcolor[HTML]{EFEFEF} 
\multirow{-3}{*}{\cellcolor[HTML]{EFEFEF}\textbf{\begin{tabular}[c]{@{}l@{}}BH-approx \\ (1275 pts)\end{tabular}}} & \textit{p} value &  & \textgreater{}0.1 & \textgreater{}0.1 & \textbf{6.90e-05} & \textbf{2.50e-23} \\ \hline
 & Median (SD) & 2.01 (0.18) & 2.02 (0.16) & 1.99 (0.17) & \textbf{1.93 (0.18)} & \textbf{1.79 (0.22)} \\ 
 & \textit{F} stats &  & \textless{}2.5 & \textless{}2.5 & \textbf{17.9} & \textbf{120.3} \\ 
\multirow{-3}{*}{\textbf{\begin{tabular}[c]{@{}l@{}}BH-approx \\ (2540 pts)\end{tabular}}} & \textit{p} value &  & \textgreater{}0.1 & \textgreater{}0.1 & \textbf{3.00e-05} & \textbf{6.30e-24} \\ \hline
\rowcolor[HTML]{C0C0C0} 
\multicolumn{7}{l}{\cellcolor[HTML]{C0C0C0}\textbf{BA35 thickness (mm), age as covariate}} \\ \hline
 & Median (SD) & 2.35 (0.16) & 2.32 (0.18) & 2.30 (0.19) & \textbf{2.19 (0.23)} & \textbf{2.06 (0.22)} \\ 
 & \textit{F} stats &  & 2.8 & 8.0 & \textbf{49.2} & \textbf{182.3} \\ 
\multirow{-3}{*}{\textbf{\begin{tabular}[c]{@{}l@{}}Non-approx\\ (1275 pts)\end{tabular}}} & \textit{p} value &  & 0.097 & 5.10e-03 & \textbf{1.50e-11} & \textbf{4.30e-33} \\ \hline
\rowcolor[HTML]{EFEFEF} 
\cellcolor[HTML]{EFEFEF} & Median (SD) & 2.36 (0.16) & 2.33 (0.17) & 2.31 (0.18) & \textbf{2.20 (0.22)} & \textbf{2.07 (0.22)} \\ 
\rowcolor[HTML]{EFEFEF} 
\cellcolor[HTML]{EFEFEF} & \textit{F} stats &  & 3.0 & 7.4 & \textbf{52.1} & \textbf{189.4} \\ 
\rowcolor[HTML]{EFEFEF} 
\multirow{-3}{*}{\cellcolor[HTML]{EFEFEF}\textbf{\begin{tabular}[c]{@{}l@{}}BH-approx\\ (1275 pts)\end{tabular}}} & \textit{p} value &  & 0.083 & 6.60e-03 & \textbf{4.40e-12} & \textbf{4.60e-34} \\ \hline
 & Median (SD) & 2.37 (0.16) & 2.34 (0.17) & \textbf{2.31 (0.18)} & \textbf{2.21 (0.22)} & \textbf{2.07 (0.22)} \\ 
 & \textit{F} stats &  & 3.2 & \textbf{10.5} & \textbf{50.8} & \textbf{190.0-} \\ 
\multirow{-3}{*}{\textbf{\begin{tabular}[c]{@{}l@{}}BH-approx\\ (2540 pts)\end{tabular}}} & \textit{p} value &  & 0.075 & \textbf{1.30e-03} & \textbf{7.80e-12} & \textbf{3.80e-34} \\ \hline
\rowcolor[HTML]{C0C0C0} 
\multicolumn{7}{l}{\cellcolor[HTML]{C0C0C0}\textbf{BA36 thickness (mm), age as covariate}} \\ \hline
 & Median (SD) & 2.40 (0.23) & 2.38 (0.23) & 2.38 (0.21) & \textbf{2.28 (0.22)} & \textbf{2.20 (0.24)} \\ 
 & \textit{F} stats &  & \textless{}2.5 & \textless{}2.5 & \textbf{20.4} & \textbf{56.8} \\ 
\multirow{-3}{*}{\textbf{\begin{tabular}[c]{@{}l@{}}Non-approx\\ (1275 pts)\end{tabular}}} & \textit{p} value &  & \textgreater{}0.1 & \textgreater{}0.1 & \textbf{9.00e-06} & \textbf{5.30e-13} \\ \hline
\rowcolor[HTML]{EFEFEF} 
\cellcolor[HTML]{EFEFEF} & Median (SD) & 2.42 (0.23) & 2.39 (0.23) & 2.40 (0.21) & \textbf{2.29 (0.22)} & \textbf{2.21 (0.23)} \\ 
\rowcolor[HTML]{EFEFEF} 
\cellcolor[HTML]{EFEFEF} & \textit{F} stats &  & \textless{}2.5 & \textless{}2.5 & \textbf{20.9} & \textbf{57.2} \\ huijiale
\rowcolor[HTML]{EFEFEF} 
\multirow{-3}{*}{\cellcolor[HTML]{EFEFEF}\textbf{\begin{tabular}[c]{@{}l@{}}BH-approx\\ (1275 pts)\end{tabular}}} & \textit{p} value &  & \textgreater{}0.1 & \textgreater{}0.1 & \textbf{7.00e-06} & \textbf{4.40e-12} \\ \hline
 & Median (SD) & 2.42 (0.23) & 2.41 (0.23) & 2.40 (0.21) & \textbf{2.30 (0.22)} & \textbf{2.22 (0.23)} \\ 
 & \textit{F} stats &  & \textless{}2.5 & \textless{}2.5 & \textbf{21.3} & \textbf{57.9} \\ 
\multirow{-3}{*}{\textbf{\begin{tabular}[c]{@{}l@{}}BH-approx\\ (2540 pts)\end{tabular}}} & \textit{p} value &  & \textgreater{}0.1 & \textgreater{}0.1 & \textbf{6.00e-06} & \textbf{3.20e-13} \\ \hline
\rowcolor[HTML]{C0C0C0} 
\multicolumn{7}{l}{\cellcolor[HTML]{C0C0C0}\textbf{PHC thickness (mm), age as covariate}} \\ \hline
 & Median (SD) & 2.14 (0.13) & 2.16 (0.16) & 2.14 (0.13) & \textbf{2.09 (0.15)} & \textbf{1.99 (0.15)} \\ 
 & \textit{F} stats &  & \textless{}2.5 & \textless{}2.5 & \textbf{9.9} & \textbf{92.2} \\ 
\multirow{-3}{*}{\textbf{\begin{tabular}[c]{@{}l@{}}Non-approx\\ (1275 pts)\end{tabular}}} & \textit{p} value &  & \textgreater{}0.1 & \textgreater{}0.1 & \textbf{1.80e-03} & \textbf{2.60e-19} \\ \hline
\rowcolor[HTML]{EFEFEF} 
\cellcolor[HTML]{EFEFEF} & Median (SD) & 2.14 (0.12) & 2.16 (0.16) & 2.14 (0.13) & \textbf{2.09 (0.14)} & \textbf{1.99 (0.15)} \\ 
\rowcolor[HTML]{EFEFEF} 
\cellcolor[HTML]{EFEFEF} & \textit{F} stats &  & \textless{}2.5 & \textless{}2.5 & \textbf{10.8} & \textbf{94.9} \\ 
\rowcolor[HTML]{EFEFEF} 
\multirow{-3}{*}{\cellcolor[HTML]{EFEFEF}\textbf{\begin{tabular}[c]{@{}l@{}}BH-approx\\ (1275 pts)\end{tabular}}} & \textit{p} value &  & \textgreater{}0.1 & \textgreater{}0.1 & \textbf{1.10e-03} & \textbf{8.90e-20} \\ \hline
 & Median (SD) & 2.15 (0.13) & 2.17 (0.16) & 2.15 (0.13) & \textbf{2.10 (0.14)} & \textbf{2.00 (0.15)} \\ 
 & \textit{F} stats &  & \textless{}2.5 & \textless{}2.5 & \textbf{13.9} & \textbf{97.6} \\ 
\multirow{-3}{*}{\textbf{\begin{tabular}[c]{@{}l@{}}BH-approx\\ (2540 pts)\end{tabular}}} & \textit{p} value &  & \textgreater{}0.1 & \textgreater{}0.1 & \textbf{2.20e-04} & \textbf{3.10e-20} \\ \hline
\end{tabular}
\end{table}

%
%
%
%